# Evaluating Supervised Learning Models for Fraud Detection: A Comparative Study of Classical and Deep Architectures on Imbalanced Transaction Data

Chao Wang[1], Chuanhao Nie[2], Yunbo Liu[3*]

[1]Department of Computer Science, Rice University, Houston, United States
[2]College of Computing, Georgia Institute of Technology, Atlanta, United States
[3]Department of Electrical and Computer Engineering, Duke University, Durham, United States
Chao Wang and Chuanhao Nie are co-first authors and contributed equally to this work

*Corresponding author: yunbo.liu954@duke.edu

**Abstract.** Fraud detection remains a critical task in high-stakes domains such as finance and e-commerce, where undetected fraudulent transactions can lead to significant economic losses. In this study, we systematically compare the performance of four supervised learning models—Logistic Regression, Random Forest, Light Gradient Boosting Machine (LightGBM), and a Gated Recurrent Unit (GRU) network—on a large-scale, highly imbalanced online transaction dataset. While ensemble methods such as Random Forest and LightGBM demonstrated superior performance in both overall and class-specific metrics, Logistic Regression offered a reliable and interpretable baseline. The GRU model showed strong recall for the minority fraud class, though at the cost of precision, highlighting a trade-off relevant for real-world deployment. Our evaluation emphasizes not only weighted averages but also per-class precision, recall, and F1-scores, providing a nuanced view of each model's effectiveness in detecting rare but consequential fraudulent activity. The findings underscore the importance of choosing models based on the specific risk tolerance and operational needs of fraud detection systems.

## 1 Introduction

Fraud continues to be a pervasive issue across industries such as banking, insurance, and healthcare, inflicting not only financial damage but also reputational harm [1]. In financial services alone, losses from fraudulent transactions reach into the billions annually, prompting a growing reliance on automated detection technologies. Traditional systems built on static rules and manual screening are increasingly inadequate—they often fail to detect emerging fraud strategies and tend to generate high false-positive rates, burdening investigative teams and delaying legitimate transactions.

In recent years, machine learning (ML) has taken center stage in the fight against fraud. Unlike rule-based systems, ML algorithms can adapt to changing behavior and uncover intricate patterns in transactional data. Among supervised approaches, models like Logistic Regression, Random Forests, and LightGBM have become standard choices due to their ease of use and strong empirical performance. However, a persistent challenge remains: class imbalance. Fraud cases typically account for only a tiny fraction of transactions, making it difficult for models to learn meaningful decision boundaries and causing traditional metrics like accuracy to overstate performance.

To address this, researchers have turned to deep learning, particularly models that can process sequential or tokenized input formats. In this work, we explore the use of a Gated Recurrent Unit (GRU) network as a representative deep architecture capable of capturing complex temporal or categorical dependencies. While deep models offer the potential to discover patterns missed by conventional methods, they often require more data, greater tuning effort, and come with trade-offs in interpretability and latency—factors that are critical in high-stakes domains like fraud prevention.

This study offers a comprehensive comparison of four supervised learning models—Logistic Regression, Random Forest, LightGBM, and GRU—applied to a large-scale, highly imbalanced online transaction dataset. Our emphasis is not only on overall model accuracy but also on each model's ability to detect fraudulent transactions specifically, which are rare but operationally vital to capture. We highlight how performance varies across models and discuss practical considerations such as false alarm rates, detection sensitivity, and resource implications.

Our contributions include:

- A comparative analysis of classical machine learning and deep learning models for fraud detection, with detailed evaluation of both weighted and minority class metrics.
- An examination of trade-offs between fraud recall and precision, underscoring the importance of model selection based on operational goals.
- A discussion of the interpretability-performance balance, illustrating where simpler models may be preferred and when more complex architectures like GRUs provide added value.

The remainder of this paper is organized as follows. Section 2 outlines the dataset, feature engineering, and modeling workflow. Section 3 presents the empirical results with both aggregate and fraud-specific evaluation. Section 4 offers a discussion on practical implications, limitations, and potential extensions. Section 5 concludes with key takeaways and directions for future research.

## 2 Methods

### 2.1 Dataset and Preprocessing

In this project, we employed the publicly available IEEE-CIS Fraud Detection dataset, which mimics a real-world e-commerce transaction environment and was published in collaboration with Vesta Corporation. The dataset contains over 590,000 anonymized online payment transactions, each labeled as fraudulent or legitimate. This dataset includes a wide range of structured features designed to mimic real-world fraud detection scenarios. It contains transactional-level details such as *TransactionAmt* (transaction amount), *ProductCD* (product type), as well as *card1* through *card6* (various card identifiers). In addition, it captures user and account metadata, including address-related fields like *addr1* and *addr2*, as well as email domains such as *P_emaildomain* and *R_emaildomain* (for email domains). The dataset also incorporates device and digital identity signals, including *DeviceType*, *DeviceInfo*, and over 30 anonymized *id_features* that reflect browser versions, operating systems, and other behavioral fingerprints. Together, these features provide a comprehensive foundation for training models to identify fraudulent transaction patterns. A binary label (*isFraud*) indicates whether a given transaction was fraudulent.

Benchmarking [2] provides standardized datasets and methodologies, which help to assess the effectiveness of different models under controlled conditions. Fraudulent transactions account for a small portion of the dataset (less than 4%), reflecting the inherent class imbalance found in real-world fraud detection problems. To ensure realistic evaluation and training integrity, we used the official train-test split from the Amazon Fraud Dataset Benchmark[3], resulting in 561,013 samples for training and 29,527 samples for testing.

In our project, the training data was further divided into 80% training and 20% validation subsets (448,810 and 112,203 samples respectively) using stratified sampling to maintain class distribution. Feature dimensionality after preprocessing was standardized at 74 columns. For preprocessing, categorical features such as *ProductCD*, *card1-card6*, *DeviceType*, and email domains were label-encoded. Timestamp variables were converted into interpretable features such as hour of day, weekday, and month. Missing values were imputed using a constant-fill strategy, and irrelevant identifiers such as transaction IDs and entity IDs were removed. All processing was applied uniformly across training, validation, and test splits.

### 2.2 Machine Learning Models

To systematically examine the trade-offs between model complexity, interpretability, and fraud detection performance under class imbalance, we evaluated three supervised learning algorithms: Logistic Regression, Tree-based Models, including Random Forest (i.e., bagging techniques), and Light Gradient Boosting Machine (LightGBM, i.e., boosting techniques). These models represent a spectrum of approaches—from transparent linear classification to sophisticated tree-based ensembles—and were selected to assess how well different algorithmic families adapt to the challenges of real-world fraud detection [4,5,6]. For each model, we conducted hyperparameter tuning using grid search over relevant parameter spaces, with the primary evaluation metric being the weighted F1 score on a held-out validation set.

*Logistic Regression*

Logistic Regression [7] serves as a baseline model with a linear classification boundary, offering high interpretability and ease of implementation. It estimates the likelihood of a transaction being fraudulent as a logistic function of a weighted sum of the input features. Although it lacks the flexibility to capture complex feature interactions or non-linear decision boundaries, its transparency is advantageous in regulated domains like finance. To account for the class imbalance in our data, we enabled the *class_weight='balanced'* option, which adjusts the model's loss function to penalize misclassification of minority class instances more heavily. We performed grid search over regularization strength (*C*), solver type (*lbfgs* or *liblinear*), and class weighting to identify the configuration that maximized F1 score on the validation set.

*Decision Trees*

Decision Trees build a hierarchy of if-then rules by recursively partitioning the data to minimize node impurity, such as using Gini index or information gain. They are capable of modeling nonlinear relationships and are highly interpretable, as the decision path for any prediction is explicitly defined. However, standalone trees are prone to overfitting, especially in high-dimensional datasets or when class distributions are skewed. In preliminary experiments, a single tree demonstrated poor generalization on the validation set, leading us to exclude it from further evaluation in favor of ensemble methods.

*Random Forest: Bagging Techniques*

Random Forest [8] is an ensemble learning method that constructs multiple decision trees using bootstrapped subsets of the data and aggregates their outputs via majority voting. This strategy reduces overfitting and enhances model stability while preserving the tree model's capacity to handle nonlinear interactions and heterogeneous feature types. We used the *class_weight='balanced'* parameter when training to ensure that the model gave appropriate emphasis to the minority (fraudulent) class. A grid search over key hyperparameters, such as tree depth, number of estimators, and minimum leaf size, was performed to optimize performance.

*Light Gradient Boosting Machine: Boosting Techniques*

LightGBM [9,10] is a high-efficiency gradient boosting framework designed for scalable learning on large, sparse datasets. Unlike Random Forest, which builds trees in parallel, LightGBM grows trees sequentially, where each new tree is trained to correct the residuals of the ensemble thus far. It employs a leaf-wise tree growth strategy and histogram-based feature binning, which accelerates training and reduces memory usage. This model handles class imbalance through its gradient-based optimization and splitting heuristics. Hyperparameter tuning was performed over key settings such as the number of estimators, learning rate, maximum tree depth, number of leaves, and minimum child samples. The selected configuration optimized performance while maintaining computational efficiency. All supervised models were trained on the labeled training set, and hyperparameters were tuned using the validation set. Oversampling of the minority class was applied to mitigate imbalance effects.

### 2.3 Deep Learning Models

To complement the classical machine learning methods [11,12,13,14], we implemented neural architectures capable of capturing complex patterns in sequence-based or tokenized categorical features: recurrent network (GRU). It is fine-tuned for binary fraud classification, with performance evaluated using the weighted F1 score.

*Gated Recurrent Unit (GRU)*

The GRU model was implemented in PyTorch and consisted of an embedding layer, a GRU layer, and a fully connected output layer. It was designed to capture sequential dependencies across tokenized categorical features. To mitigate overfitting and handle class imbalance, dropout regularization was applied and weighted cross-entropy loss was used, with class weights derived from the training distribution. Hyperparameter tuning was performed through random sampling across 10 iterations, exploring embedding dimensions (150–250), hidden units (256–768), learning rates ($10^{-4}$-$10^{-3}$), and training epochs (5–10). Each sampled configuration was trained using the Adam optimizer and evaluated on the validation set, with the weighted F1-score serving as the primary selection criterion.

Both models demonstrated the ability to learn from tokenized categorical inputs, offering complementary strengths to tree-based classifiers in fraud detection under imbalanced conditions.

### 2.4 Evaluation Metrics

Given the severe class imbalance inherent in fraud detection, overall accuracy is often misleading, as a model biased toward the majority class can achieve high accuracy while failing to detect fraudulent cases. To provide a more meaningful evaluation, we assessed all models using four standard metrics: precision, recall, F1-score, and Area Under the Receiver Operating Characteristic Curve (AUC-ROC).

Precision measures the proportion of predicted fraud cases that are actually fraudulent, while recall indicates the proportion of true fraud cases successfully identified. In the context of this study, recall is particularly important, as failing to detect a fraudulent transaction (a false negative) can carry high financial and security risks. At the same time, precision remains operationally relevant, as a high false positive rate can overwhelm fraud analysts and reduce trust in the system. To balance these competing priorities, we relied on the F1-score as the primary model selection criterion, as it synthesizes both precision and recall into a single, interpretable metric. AUC-ROC was used to assess the overall discriminative ability of each model, capturing performance across varying decision thresholds.

This evaluation framework was applied consistently across both traditional and deep learning models, ensuring comparability under imbalanced classification conditions.

## 3 Results

### 3.1 Hyperparameter Selection

For each supervised learning model, we conducted systematic hyperparameter tuning to identify configurations that optimized performance under class imbalance. Model-specific hyperparameters were selected based on validation weighted F1-score, which balances precision and recall—two metrics critical in the context of fraud detection.

Grid search was employed for machine learning models as indicated in Section II. For Logistic Regression, we varied the regularization strength (C), solver type (liblinear vs. lbfgs), and class weighting (None vs. balanced). For Random Forest, the search space included the number of trees, maximum depth, and minimum leaf size, in addition to class weighting. LightGBM parameters were selected by adjusting the number of estimators, learning rate, tree depth, and number of leaves.

In addition to tuning classical models, we conducted random hyperparameter search for the deep learning architectures to optimize their performance under the same evaluation framework. Due to higher computational cost, a smaller number of configurations was explored for each model, with parameter ranges tailored to the architecture type. The optimal configuration selected from the GRU random search consisted of an embedding dimension of 184, hidden layer size of 502, a learning rate of $1.47 \times 10^{-4}$, and 8 training epochs. This setup achieved the best validation F1-score during tuning and was used for final evaluation on the test set.

The optimal settings for each model are summarized in Table 1. This tuning process ensured fair comparison across models and robust evaluation under the skewed class distribution typical of fraud detection tasks.

**Table 1.** Optimal Hyperparameter Configurations

| Model | Key Hyperparameters |
|---|---|
| Logistic Regression | C=100, solver=liblinear, penalty=l2, class_weight=None |
| Random Forest | n_estimators=200, max_depth=None, min_samples_leaf=4, class_weight=Balanced |
| LightGBM | n_estimators=100, lr=0.1, max_depth=10, num_leaves=50, class_weight=None |
| GRU | Embedding dimension = 184; Hidden units = 502; Learning rate = $1.47 \times 10^{-4}$; Epochs = 8 |

### 3.2 Model Performance

Table 2 summarizes the weighted average precision, recall, and F1-score for each model on the held-out test set, alongside AUC and 95% bootstrap confidence intervals. These metrics reflect aggregate performance across both classes, with weighting based on class support.

**Table 2.** Overall Model Performance

| Model | Weighted Average | | | AUC |
|---|---|---|---|---|
| | Precision | Recall | F1-Score | |
| Logistic Regression | 0.95 | 0.96 | 0.95 | 0.82 (0.81, 0.84) |
| Random Forest | 0.97 | 0.97 | 0.97 | 0.91 (0.90, 0.92) |
| LightGBM | 0.97 | 0.97 | 0.97 | 0.90 (0.89, 0.91) |
| GRU | 0.95 | 0.92 | 0.93 | 0.84 (0.83, 0.86) |

Random Forest and LightGBM achieved the highest overall scores, each recording a weighted F1-score of 0.97 and AUCs of 0.91 and 0.90, respectively. Their strong performance is attributed to their capacity to model complex feature interactions and adapt well to structured, imbalanced data. Logistic Regression, while simpler and linear, performed competitively with a weighted F1-score of 0.95 and an AUC of 0.82, offering an interpretable and computationally efficient baseline.

The GRU model, a recurrent neural network trained on tokenized categorical sequences, achieved a weighted F1-score of 0.937 and an AUC of 0.84. Despite its relatively compact architecture and smaller hyperparameter tuning budget, GRU demonstrated solid generalization across classes and competitive overall performance.

All models showed high weighted precision and recall, largely due to their accuracy on the majority class (non-fraud). However, these aggregate metrics can mask performance disparities on the minority class, which is critical in fraud detection.

Table 3 provides a focused comparison of each model's precision, recall, and F1-score for the minority (fraud) class, which comprised only 3.96% of the test data. These metrics offer a clearer view of how well each model handled the core challenge of fraud detection.

**Table 3.** Minority Class (Fraud) Performance

| Model | Precision (Fraud) | Recall (Fraud) | F1-Score (Fraud) |
|---|---|---|---|
| Logistic Regression | 0.69 | 0.10 | 0.18 |
| Random Forest | 0.79 | 0.40 | 0.53 |
| LightGBM | 0.76 | 0.38 | 0.50 |
| GRU | 0.28 | 0.54 | 0.37 |

While GRU achieved the highest recall (0.54), capturing more actual fraud cases, it did so at the cost of precision (0.28), indicating a relatively high false positive rate. Logistic Regression, by contrast, maintained high precision (0.69) but detected only 10% of true frauds, limiting its practical utility in fraud detection. Random Forest and LightGBM offered a more favorable balance between precision and recall, each achieving F1-scores near 0.50 on the fraud class.

This contrast highlights the limitations of relying solely on weighted metrics in imbalanced classification. From a fraud detection standpoint, where missing fraudulent transactions may carry significant consequences, models like GRU or Random Forest may be preferred over more conservative classifiers like Logistic Regression.

## 4 Discussion

### 4.1 Interpretation of Findings

This study systematically evaluated four supervised models—Logistic Regression, Random Forest, LightGBM, and a GRU-based neural network—on a highly imbalanced fraud detection dataset. The results underscore a clear performance gap between simple linear models and more complex architectures. Both Random Forest and LightGBM achieved top-tier performance in terms of overall weighted metrics and minority class detection, confirming their ability to capture non-linear interactions and detect rare fraud patterns.

Logistic Regression, while computationally efficient and interpretable, struggled with the imbalance, achieving high precision but extremely low recall on the fraud class. This indicates that it correctly identified a small number of fraud cases with confidence, but missed the majority. Its poor recall underscores the difficulty linear models face when fraudulent behavior does not manifest as linearly separable patterns.

GRU, in contrast, demonstrated stronger fraud recall (0.54) than Logistic Regression and comparable F1-score to tree-based models, despite being trained on tokenized inputs with less handcrafted feature engineering. This highlights the GRU's ability to learn sequential or contextual representations from categorical fields. However, its relatively low fraud precision (0.28) suggests a higher false positive rate—an important trade-off to consider depending on operational constraints.

All models demonstrated high weighted F1-scores, largely driven by strong performance on the majority class. This reinforces the importance of looking beyond aggregate metrics and evaluating class-specific performance when working with imbalanced data. In our case, fraud recall was particularly prioritized, given the high cost of undetected fraudulent transactions.

### 4.2 Practical Implications

Our findings offer several practical insights for deploying fraud detection systems:

- Logistic Regression remains a viable option in regulated environments or audit-sensitive pipelines where model transparency is critical. However, it is best suited for environments with well-engineered features and relatively balanced data.
- Random Forest and LightGBM demonstrated strong out-of-the-box performance, scalability, and robustness under class imbalance. LightGBM, in particular, offers fast training and inference, making it well-suited for real-time fraud monitoring systems.
- The GRU model shows promise in learning from minimally processed, tokenized data. This can be advantageous in systems where raw categorical sequences are abundant but manually engineered features are sparse [15]. With further tuning or hybridization (e.g., attention mechanisms), its practical utility may improve.

Given the trade-offs between precision and recall observed across all models, threshold adjustment, cost-sensitive training, or ensemble calibration strategies may be necessary to align model behavior with specific business objectives.

### 4.3 Limitations and Future Works

This study has several limitations that warrant further exploration:

- The dataset, while rich and realistic, is public and partially anonymized, which may not fully reflect the diversity or evolving nature of fraud in proprietary financial systems.
- The deep learning component was limited to GRU. Future work should investigate more advanced architectures such as transformer-based models (e.g., ALBERT, BERT), or hybrid models that integrate structured and unstructured features [16,17]. Recent advances such as FutureSightDrive [18] have demonstrated the potential of combining spatial and temporal reasoning with visual transformers, suggesting promising directions for applying similar CoT-style architectures in high-stakes environments like fraud detection [19]. Similarly, NeRF [20], OFFSET [21] and PAIR [22] provide insights into segmentation and disentanglement strategies that may inspire fraud detection under compositional semantics.
- While our evaluation focused on standard classification metrics, we did not incorporate cost-sensitive learning or business-impact-weighted metrics, which are critical in production environments where false positives and false negatives carry asymmetric costs.
- All models were trained in a static, batch-learning setting. In practice, fraud patterns change dynamically. Future research should consider online learning, continual learning, or adaptive drift-handling frameworks to better simulate real-time deployment scenarios. Additionally, recent progress in model unlearning techniques provides a promising pathway for adapting fraud detection systems to evolving transaction patterns without full retraining. Approaches such as error-decomposition-based unlearning [23] and post-training attribute-level forgetting [24] may allow fraud models to remove outdated behavior patterns while preserving core detection capabilities.

## 5 Conclusion

In this study, we conducted a comprehensive evaluation of four supervised learning models—Logistic Regression, Random Forest, LightGBM, and a GRU-based neural network—for fraud detection on a highly imbalanced dataset. The results highlighted the superior ability of tree-based models and recurrent neural networks to handle class imbalance and capture complex patterns indicative of fraud. While Logistic Regression offered interpretability and computational efficiency, it significantly underperformed on recall for the minority class, limiting its practical utility in high-risk settings.

LightGBM and Random Forest emerged as robust, high-performing solutions, particularly suited for scenarios demanding scalability and low latency. Meanwhile, the GRU-based model showed promise in learning from raw, tokenized categorical sequences, suggesting a valuable path forward for systems with limited feature engineering.

Our findings reinforce the necessity of evaluating models not only on aggregate performance but also on class-specific metrics, especially in high-stakes domains like fraud detection where recall for the positive class (fraud) is critical. Furthermore, the observed trade-offs between precision and recall across models underline the importance of threshold tuning and cost-sensitive learning strategies in real-world deployment.